\documentclass[final,onefignum,onetabnum]{siuro210301}

\usepackage{lipsum}
\usepackage{amsfonts}
\usepackage{graphicx}
\usepackage{epstopdf}
\usepackage{algorithmic}
\usepackage{algorithm}
\usepackage{subcaption}
\usepackage{caption}
\usepackage [english]{babel}
\usepackage [autostyle, english = american]{csquotes}
\MakeOuterQuote{"}
\usepackage[section]{placeins}
\usepackage{float}

\newcommand{\eps}{\varepsilon}

\ifpdf
  \DeclareGraphicsExtensions{.eps,.pdf,.png,.jpg}
\else
  \DeclareGraphicsExtensions{.eps}
\fi

\usepackage{enumitem}
\setlist[enumerate]{leftmargin=.5in}
\setlist[itemize]{leftmargin=.5in}
\newsiamremark{remark}{Remark}
\newsiamremark{hypothesis}{Hypothesis}
\Crefname{hypothesis}{Hypothesis}{Hypotheses}
\newsiamthm{claim}{Claim}

\headers{Generative Modeling with Diffusion}{Justin Le}

\title{Generative Modeling with Diffusion}

\author{Justin Le\thanks{Arizona State University, Tempe, AZ 
  (\email{jdle4@asu.edu}).}}

\dedication{\small\textit{Project advisors: Sebastien Motsch\thanks{Arizona State University, Tempe, AZ 
  (\email{smotsch@asu.edu}).} and Johannes Brust\thanks{Arizona State University, Tempe, AZ 
  (\email{jjbrust@asu.edu}).}}}

\usepackage{amsopn}

\ifpdf
\hypersetup{
  pdftitle={Generative Modeling with Diffusion},
  pdfauthor={Justin Le}
}
\fi

\begin{document}

\maketitle

\begin{abstract}
  We provide an overview of the diffusion model as a method to generate new samples.\ Generative models have been recently adopted for tasks such as art generation (Stable Diffusion, Dall-E) and text generation (ChatGPT). Diffusion models in particular apply noise to sample data and then "reverse" this noising process to generate new samples. We will formally define these noising and denoising processes, then present algorithms to train and generate with a diffusion model. Afterward, we will explore a potential application of diffusion models in improving classifier performance on imbalanced data.
\end{abstract}

\section{Introduction}
\label{sec:intro}
\begin{figure}[H]
    \centering
    \includegraphics[width=\textwidth]{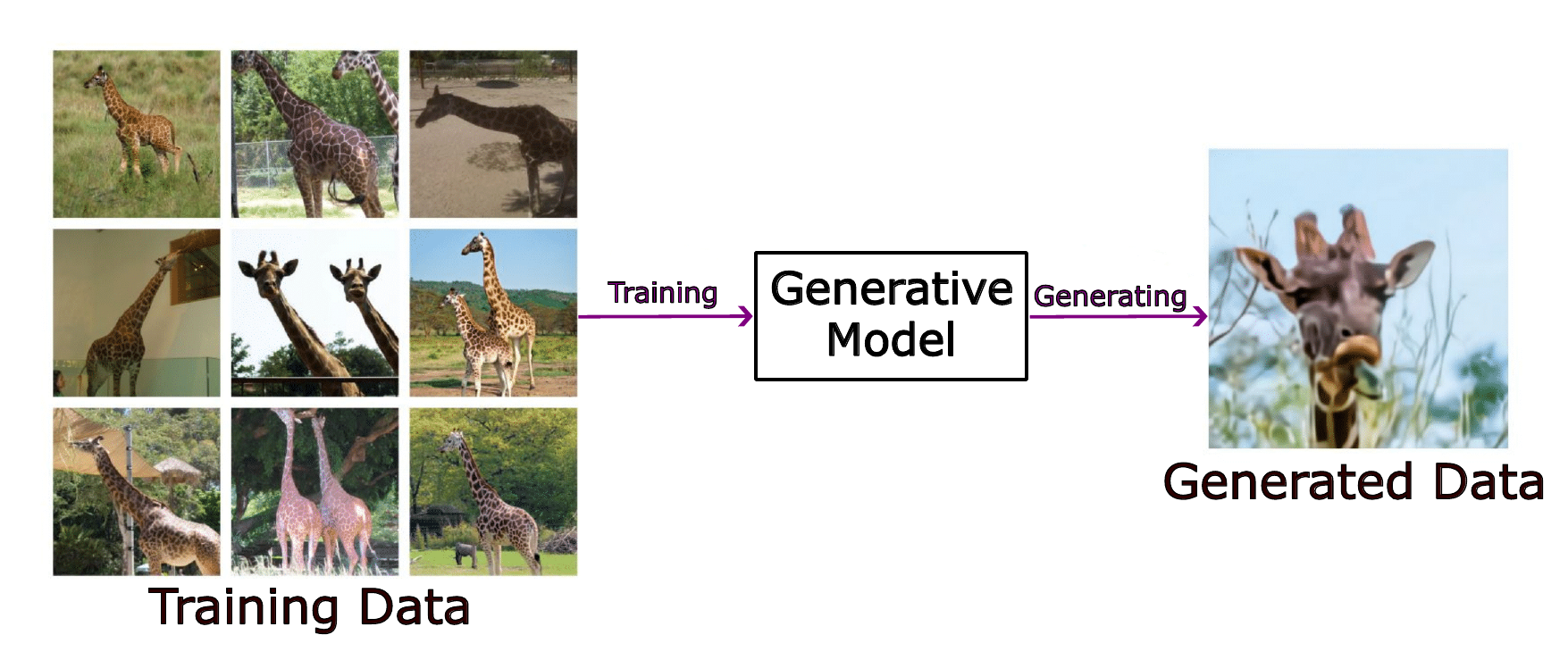}
    \caption{A sketch of the pipeline for image generation. On the left are real images of giraffes, and on the right is an image of a giraffe generated by Dall-E (found in \cite{dalle_paper}).}
    \label{fig:gen_demo}
\end{figure}

\setcounter{equation}{0}
Generative models create new data instances. Recently, generative models have seen applications in image generation \cite{dalle_paper} and text generation \cite{GPT_paper}. In this article, we are interested in generative models that create data instances that "mimic" some given sample data. Doing so will allow us to sample arbitrarily many points from the distribution underlying the original sample data. \Cref{fig:gen_demo} depicts this process applied to image generation. In a broad sense, the model receives a sample of training images, which "learns" the structure of the images, then uses that knowledge to create new images that mimic the original sample. Common architectures used for generative models include Generative Adversarial Networks \cite{GAN, gan_intro}, Variational Autoencoders \cite{VAE}, and Transformers \cite {Transformer, transgan}. For this article, we will focus on discussing the mechanics of diffusion models \cite{ddpm_ho, ddm_zhu, diffusion_thermo}. In particular, we seek to give a formal overview of how diffusion models are trained and how they generate new samples.
\\~\\
In \Cref{sec:classifier_section}, we explore an application of diffusion models for improving classifier performance (this is in contrast to the typical application of diffusion models as image generators \cite{ddpm_ho}). To this end, we will focus on a dataset of credit card transactions, of which an exceedingly small minority are fraudulent, and the rest are legitimate. Our aim is to use diffusion models to generate data mimicking the fraudulent transactions, then augmenting the training data with this generated data before training a classifier.

\subsection{Diffusion Models}
\label{sec:intro_diffusion}
Diffusion models iteratively apply noise to data until the initial distribution of the data is unrecognizable, and then, after learning this "noising phase", reverse the noising to recover the initial sample. We will refer to this noising phase as the \textit{forward process} and this denoising phase the \textit{reverse process}. More specifically, the forward process will transform the sample data into a standard normal distribution (that is, normal with mean 0 and unit variance). Then, to generate new data, we sample \textit{new} points from a standard normal distribution and apply the reverse process to recover an approximation of the sample data. This scheme will allow us to generate arbitrarily many points that mimic the initial sample. 
\\
In addition to converging to a standard normal, we would also like the noising process to be continuous, converge quickly, and have randomness enforced at each time step. The randomness in the noising process is key, as this is what allows the generated data to be slightly different from the original sample when we denoise the data. To formally define the noising process, we will present the Ornstein-Uhlenbeck equation in \Cref{sec:sde_intro}, which is a type of Stochastic Differential Equation (SDE). The solution to this equation will satisfy the above properties and will therefore be used to define the forward process. After defining the forward process, we can then deduce the reverse process and use this to generate samples.

\section{Theory of Diffusion Models}
\label{sec:sde_intro}
Our first step towards developing diffusion models is to formally define the forward process. We achieve this with the following SDE.
\subsection{The Ornstein-Uhlenbeck Equation}
\label{sec:forward}
We consider the Ornstein-Uhlenbeck equation in $\mathbb{R}^d$ given by
\begin{equation}
    \label{eq:Ornstein-Uhlenbeck}
        \begin{cases}
        \mathrm{d}\textbf{X}_t & = -\textbf{X}_t \, \mathrm{d}t + \sqrt{2} \, \mathrm{d}\textbf{B}_t \\
        \textbf{X}(0) & = \textbf{X}_0
    \end{cases}.
\end{equation}
In contrast to an ODE, whose solution is a function from $[0, \infty)$ to $\mathbb{R}^d$, the solution $\textbf{X}_t$ to $\eqref{eq:Ornstein-Uhlenbeck}$ is a stochastic process, which can be thought of as a function from $[0, \infty)$ into the space of random variables on $\mathbb{R}^d$. Likewise, the initial condition $\textbf{X}_0$ is our original sample rather than a point in $\mathbb{R}^d$. The overarching goal of a diffusion model is to generate samples from the underlying distribution of $\textbf{X}_0$.
\\~\\
The term $-\textbf{X}_t\mathrm{d}t$ is known as the \textit{drift} and is the non-stochastic element of the SDE. Intuitively, this drift term contributes exponential decay, as it is known that the ODE
\begin{align*}
    \begin{cases}
        \mathrm{d}\textbf{X}_t & = -\textbf{X}_t \, \mathrm{d}t \\
        \textbf{X}(0) & = x_0
    \end{cases}
\end{align*}
has as its solution
\begin{align*}
    \textbf{X}_t = x_0 e^{-t}.
\end{align*}
The stochastic process $\textbf{B}_t \sim \mathcal{N}(\textbf{0}, t\textbf{I})$ is responsible for enforcing randomness in this SDE. We call $\textbf{B}_t$ a \textit{Wiener process} (or \textit{Brownian motion}). A Wiener process can be thought of as modeling particles dispersing in space over time.
\\~\\
We begin by stating the well-known solution to the Ornstein-Uhlenbeck equation. 
\begin{proposition}
    \label{Prop_OUE_solution}
    The solution to \eqref{eq:Ornstein-Uhlenbeck} is given by
\begin{equation}
    \label{eq:Ornstein-Uhlenbeck_Solution_Integral}
    \mathbf{X}_t = e^{-t}\mathbf{X}_0 + + \sqrt{2}\int_0^t e^{-(t-s)} \, \mathrm{d}\mathbf{B}_s,
\end{equation}
and the distribution of $\mathbf{X}_t$ satisfies
\begin{equation}
    \label{eq:Ornstein-Uhlenbeck_Solution_Unsimplified}
        \mathbf{X}_t = e^{-t}\mathbf{X}_0 + \sqrt{1 - e^{-2t}} \, \mathbf{Z}, \\
\end{equation}
where $\mathbf{Z} \sim \mathcal{N}(\mathbf{0}, \mathbf{I})$.
\end{proposition}
This solution $\textbf{X}_t$ is the forward process described in \Cref{sec:intro_diffusion}. See \Cref{sec:forward_solution} and \cite{evans} for more information of how this solution is obtained. 
\\~\\
\noindent To condense our notation, we will denote 
\begin{align}
\gamma_t = \gamma(t) &:= e^{-t} \label{eq:gamma_def} \\
\beta_t = \beta(t) &:= \sqrt{1 - e^{-2t}}  \label{eq:beta_def}
\end{align}
for all $t > 0$. 
Then, \eqref{eq:Ornstein-Uhlenbeck_Solution_Unsimplified} becomes
\begin{equation}
    \label{eq:Forward_Solution}
        \textbf{X}_t = \gamma_t\textbf{X}_0 + \beta_t \textbf{Z}. \\
\end{equation}
In particular, $\textbf{X}_t \sim \mathcal{N}(\gamma_t\textbf{X}_0, \beta_t^2\textbf{I})$ for all $t \in \mathbb{R}$. Since $\lim\limits_{t \to \infty} \gamma_t = 0$ and $\lim\limits_{t \to \infty} \beta_t = 1$, we have that $\textbf{X}_t$ converges in distribution to a standard normal as $t \to \infty$. Moreover, $\gamma_t$ and $\beta_t$ are continuous and converge exponentially to their respective limits, meaning the forward process has the desired properties of randomness, rapid convergence to a standard normal, and continuity.

\subsection{Discretizing the Time Domain}
\label{sec:discretize}
\begin{figure}[H]
\centering
\begin{subfigure}{.5\textwidth}
  \centering
  \includegraphics[width=1.05\linewidth]{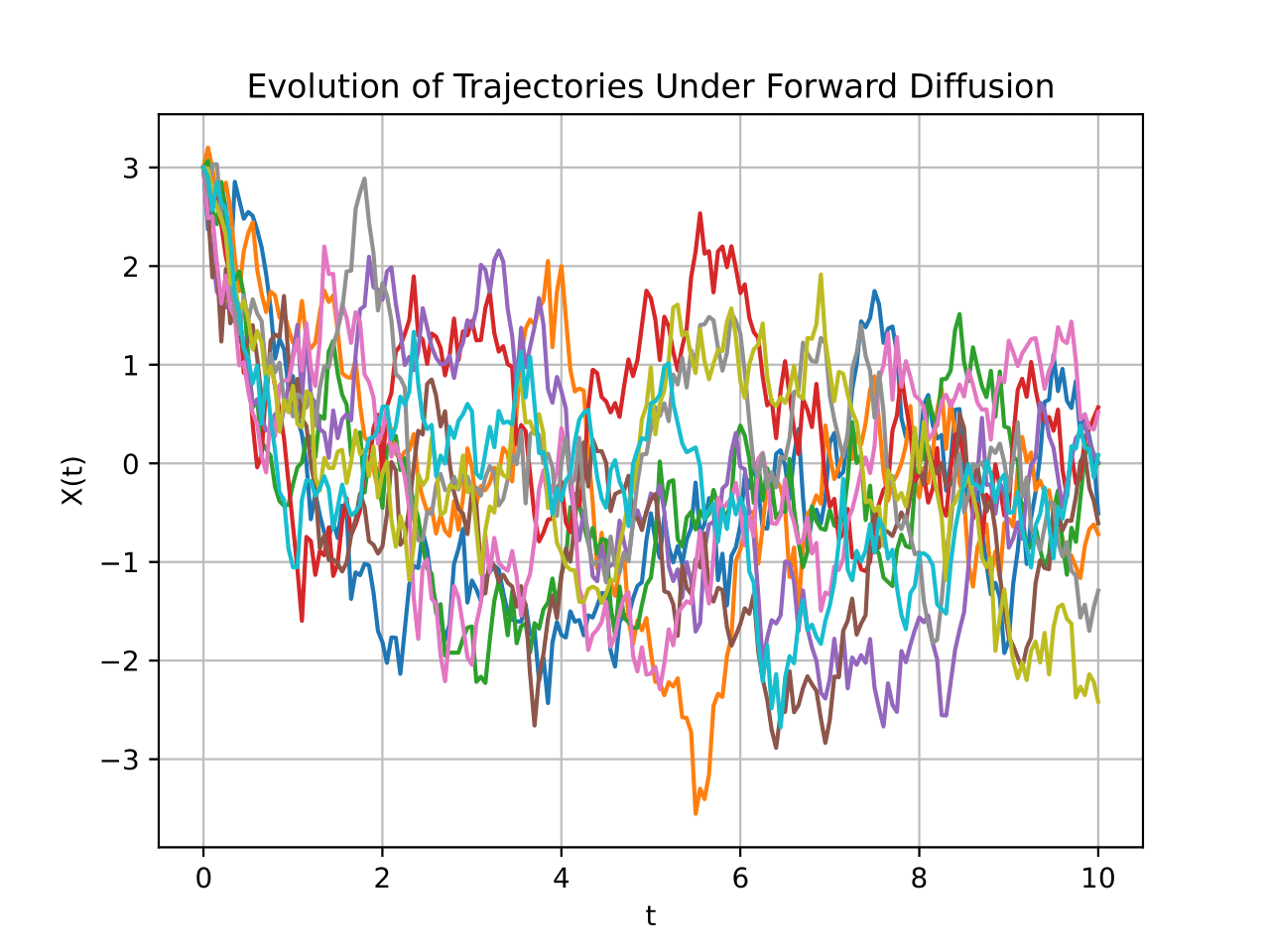}
  \caption{10 forward trajectories}
\end{subfigure}%
\begin{subfigure}{.5\textwidth}
  \centering
  \includegraphics[width=1.05\linewidth]{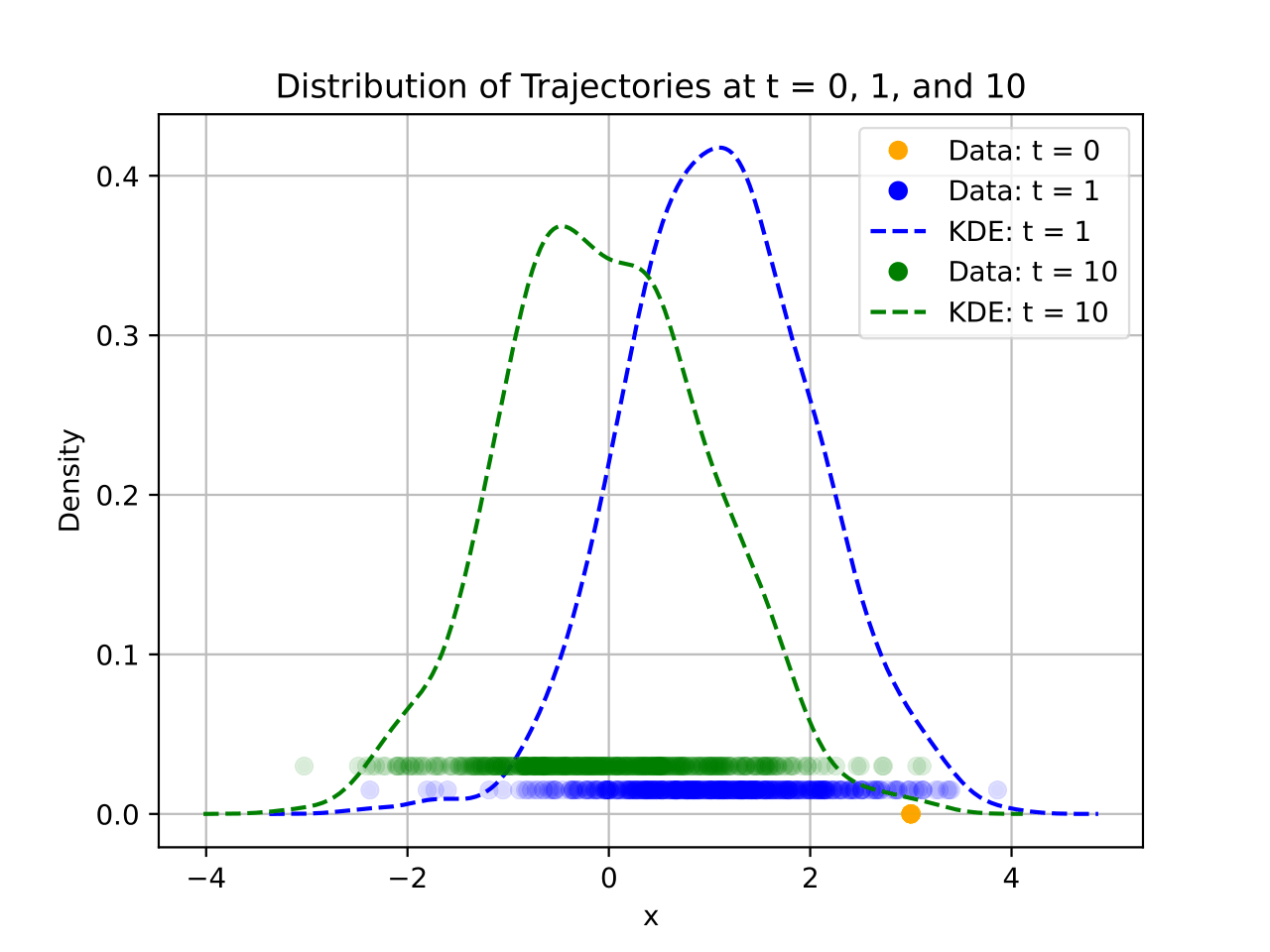}
  \caption{KDE (500 trajectories)}
\end{subfigure}
\caption{Trajectories and KDE with $\textbf{X}_0 = 3$ under the forward process.}
\label{fig:forward}
\end{figure}

We now consider a discretization of the time domain to simulate solutions to \eqref{eq:Ornstein-Uhlenbeck}. Fixing $N \in \mathbb{N}$, let $\{\Delta t_n\}_{n=1}^{N}$ be a finite sequence of positive real numbers (we discuss a more detailed scheme to define the $\Delta t_n$ in \Cref{sec:time_discretization}). Then, define $t_0 = 0$ and $t_n = \sum\limits_{k=1}^{n} \Delta t_k$ for $n = 1, \ldots, N$. In other words, the $t_n$ represent our discrete time values while the $\Delta t_n$ describe the steps in time. We denote
\begin{align*}
    \textbf{X}_{n} & = \textbf{X}_{t_n}, \quad 
    \gamma_n = \gamma_{t_n}, \quad
    \beta_n = \beta_{t_n}.
\end{align*}
Using this discretization, we immediately obtain a recursive formula for computing $\mathbf{X}_n$.
\begin{corollary}
    \label{cor_recursive}
    For all $n = 0, \ldots, N-1$, the distribution of $\mathbf{X}_{n+1}$ satisfies
    \begin{equation}
    \label{eq:Forward_Solution_Recursive}
        \mathbf{X}_{n+1} = \gamma(\Delta t_{n+1}) \mathbf{X}_{n} + \beta(\Delta t_{n+1}) \mathbf{Z}_{n+1}, \\
\end{equation}
where $\mathbf{Z}_1, \ldots, \mathbf{Z}_N \sim \mathcal{N}(\mathbf{0}, \mathbf{I})$ are i.i.d.\ standard normal random variables.
\end{corollary}
This result is obtained in the same fashion as \eqref{eq:Forward_Solution} -- see \Cref{sec:forward_solution} for more details.
\\~\\
To simulate the forward process on sample data, we take $x_0 \in \mathbb{R}^d$ and iteratively apply  \eqref{eq:Forward_Solution_Recursive} with $x_0$ in place of $\textbf{X}_0$ and $\eps_{n} \in \mathbb{R}^d$ sampled from a standard normal distribution in place of $\textbf{Z}_n$. Note that the $\eps_n$ are sampled independently in each iteration.  \Cref{fig:forward} shows the trajectories and kernel density estimate (KDE) under \eqref{eq:Forward_Solution_Recursive} with $\textbf{X}_0$ being sampled from a Dirac mass centered at 3. Importantly, despite the fact that all 10 trajectories begin at the same position, each of the trajectories split off into different directions because of the randomness enforced in  \eqref{eq:Forward_Solution_Recursive}. Moreover, as $t$ increases, the distribution of $\textbf{X}_t$ begins to resemble that of a standard normal.

\subsection{Reverse Diffusion}
\begin{figure}[H]
        \includegraphics[width=.95\textwidth]{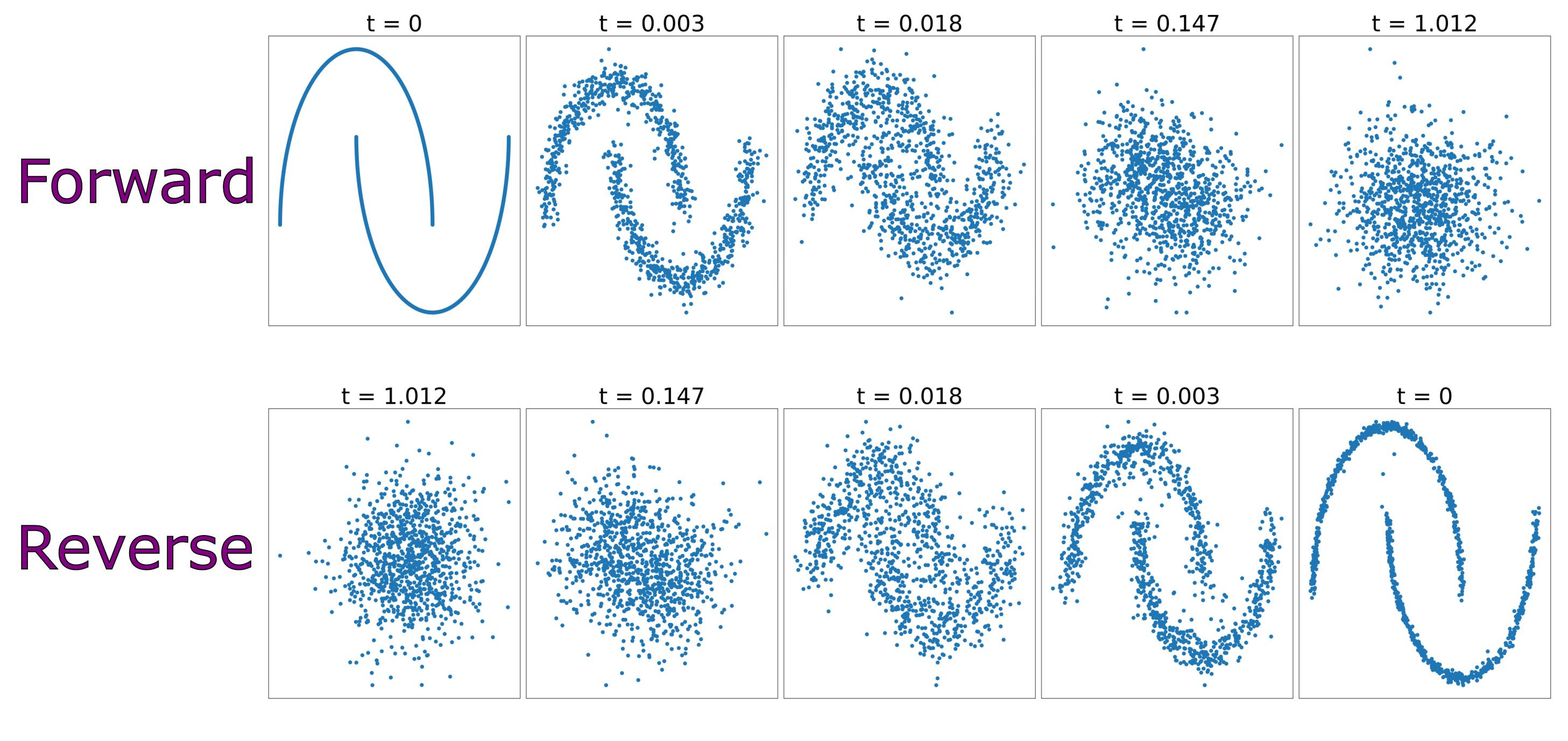}
        \caption{A timeline of the forward process (top) and the reverse process (bottom) applied to the Moons dataset from Scikit-Learn \cite{scikit-learn}. The original sample is in the top-left corner and the generated data is in the bottom-right corner.}
        \label{fig:diffusion_moons_demo}
\end{figure}

We have seen that the forward diffusion process takes any initial distribution and transforms it into a standard normal. We now ask whether or not the forward diffusion process can be reversed. That is, starting from data sampled from a standard normal distribution, we want to know if the forward diffusion process can be applied in reverse to obtain data that mimics the original sample. In \Cref{fig:diffusion_moons_demo}, we demonstrate these forward and reverse processes applied to a simple dataset in $\mathbb{R}^2$. Once again, the forward process begins with the original sample data and iteratively applies noise. To generate a new sample with the reverse process, we begin by sampling \textit{new} points from a standard normal and iteratively denoising the data. The resulting "denoised" data resembles the original sample, but is not exactly the same. Like with the forward process, the reverse process does obey an SDE (this is explored in \cite{thiery}). In this paper, we are not interested in this reverse SDE -- instead, we will deduce the reverse process by leveraging the discrete time steps we discussed in \Cref{sec:discretize}. Doing so will give us a discretized version of the reverse process.

\begin{figure}[H]
        \centering
        \hspace{-.5cm}
        \includegraphics[width=.925\textwidth]{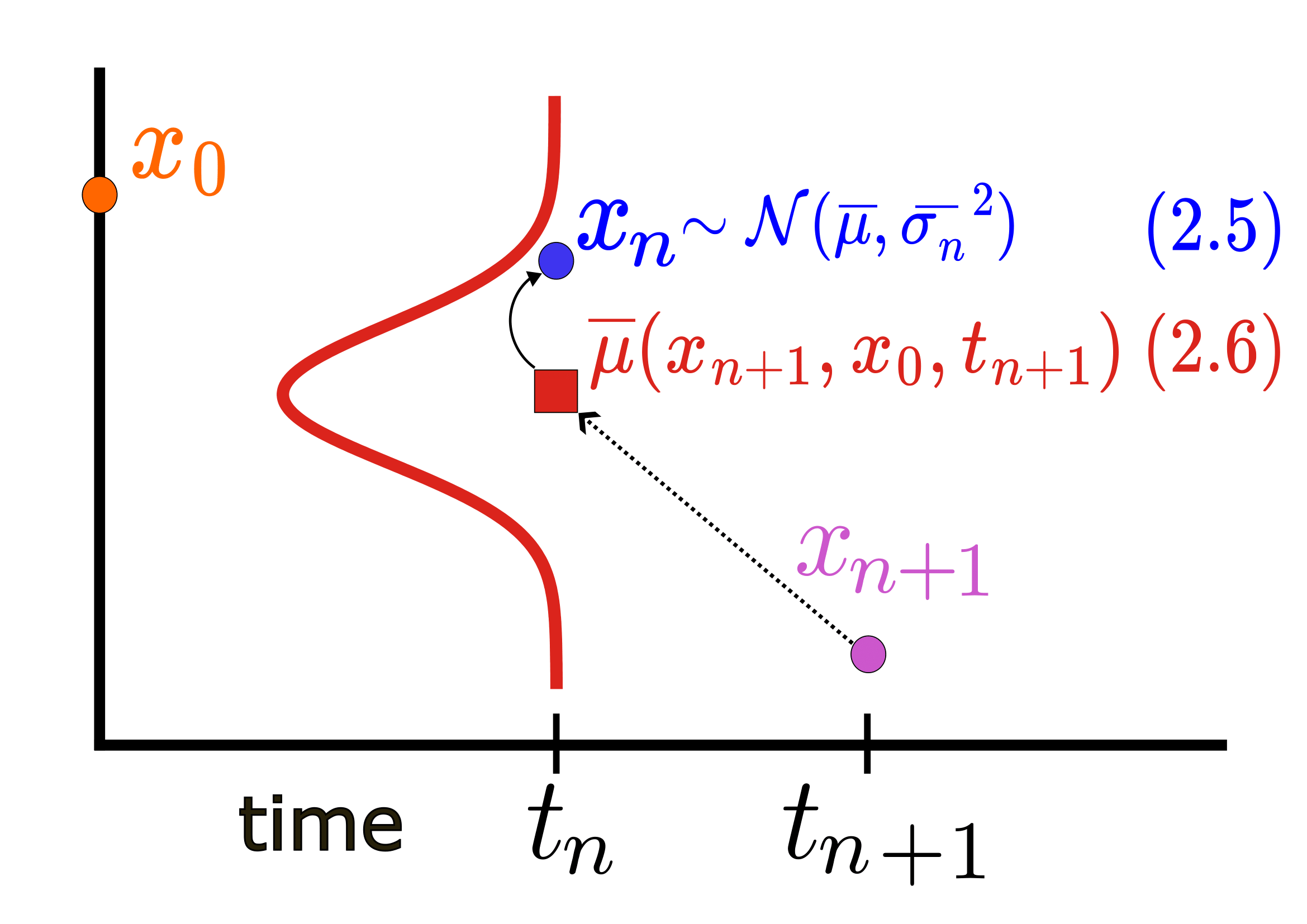}
        \caption{A sketch of the discretized reverse diffusion process with the initial condition $x_0$ known. Starting at $x_{n+1}$, we compute $\overline{\mu}$ with  \eqref{true_mu_equation}, then compute $\overline{\sigma}_n^2$ with $\eqref{std_equation}$, and finally sample $x_n \sim \mathcal{N}(\overline{\mu}, \overline{\sigma}_n^2$) as in \eqref{density_equation}.}
        \label{fig:reverse_sketch}
\end{figure}

\vspace{-.25cm}\noindent{Recall} that the discretized forward diffusion process gives a finite sequence of random variables $\textbf{X}_0, \textbf{X}_1, \ldots, \textbf{X}_N$. We will assume $N$ and $t_N$ are chosen such that $\mathbf{X}_N$ is approximately standard normal for the given initial condition $\textbf{X}_0$. Ideally, we would want to determine the conditional density $\rho(x_n \, \vert x_{n+1})$ (that is, given the future position of a trajectory, we want to determine the distribution of its present position). However, deducing this conditional density is an ill-posed problem. This is because \textit{every} initial condition converges to a standard normal under the forward diffusion process. Therefore, if we start from $\textbf{X}_N \sim \mathcal{N}(\textbf{0}, \textbf{I})$ and have no information on the initial condition $\textbf{X}_0$, the conditional density $\rho(x_{N-1} \, \vert \, x_N)$ is not unique. 
\\~\\
Instead, we will look to deduce $\rho(x_n \, \vert \, x_{n+1}, x_0)$ -- that is, given the future \textit{and initial} position of a trajectory, we want to determine the distribution of its present position. Starting from $\textbf{X}_N$ and iteratively computing the conditional density of the previous time step will result in an approximation of $\rho(x_0)$.
The following proposition addresses how this conditional density is obtained:
\begin{proposition}
    \label{prop:reverse_prop}
    Let $1 \leq n \leq N-1$ and suppose $\mathbf{X}_0 = x_0$ and $\mathbf{X}_{n+1} = x_{n+1}$. Then,
    \begin{equation}
        \label{density_equation}
        \rho(x_n \, \vert \, x_{n+1}, x_0) \sim \mathcal{N}(\overline{\mu}, \overline{\sigma}_n^2),
    \end{equation}
    with $\overline{\mu} = \overline{\mu}(x_{n+1}, x_0, t_{n+1})$ and $\overline{\sigma}_n^2$ defined as
    \begin{equation}
    \label{true_mu_equation}
        \overline{\mu}  = \frac{\gamma(\Delta t_{n+1})\beta_n^2}{\beta_{n+1}^2}\cdot x_{n+1} \, + \, \frac{\gamma_n \beta(\Delta t_{n+1})^2}{\beta_{n+1}^2} \cdot x_0
    \end{equation}
    \begin{equation}
        \label{std_equation}
        \overline{\sigma}^2_n  = \frac{\beta(\Delta t_{n+1})^2\beta_n^2 }{\beta_{n+1}^2}.
    \end{equation}
\end{proposition}
See \Cref{sec:reverse_proof} for a detailed proof of this result.
\\~\\
With this proposition, we can determine the distribution of $\textbf{X}_n$ if we know the values taken on by
$\textbf{X}_0$ and $\textbf{X}_{n+1}$. In fact, this proposition tells us that the discretized reverse diffusion process is nothing more than computing a mean and variance, then sampling from a normal distribution -- \Cref{fig:reverse_sketch} shows a sketch of this procedure. This result can also be extended to $n = 0$. In particular, if $n = 0$, then \eqref{true_mu_equation} and \eqref{std_equation} yield $\overline{\mu} = x_0$ and $\overline{\sigma}_n^2 = 0$. This is consistent with the fact that the conditional distribution of $\textbf{X}_0$ given $\textbf{X}_0 = x_0$ is $x_0$ with probability 1.
What's more, the discretized reverse diffusion will tend towards $x_0$ as $n$ tends to 0, confirming that this reverse diffusion process does indeed "reverse" the forward process outlined in \Cref{sec:forward}.
\\~\\
Importantly, the variance term $\overline{\sigma}^2_n$ only depends on the iteration $n$ (i.e., there is no dependence on the position of the trajectory). Because we control these iterations, the variance of the discretized reverse diffusion process is always known. On the other hand, the mean $\overline{\mu}$ depends on both $x_{n+1}$ and $x_0$. In other words, $\overline{\mu}$ is the only quantity of interest in the reverse diffusion process that encodes information from the distribution of $\textbf{X}_0$.
\section{Building a Diffusion Model} \label{training_section}
When generating data with the reverse diffusion process, we must begin by sampling points from a standard normal. As a result, when applying the reverse diffusion process to these points, the initial condition $x_0$ (i.e., the value taken on by $\textbf{X}_0$) is not known, meaning $\overline{\mu}$ cannot be directly computed through \eqref{true_mu_equation}. To circumvent this, we will construct a scheme to estimate $\overline{\mu}$ while only knowing $x_{n+1}$ and $t_{n+1}$. For this, we employ techniques from machine learning to "learn" the value of $\overline{\mu}$ from the original sample. Letting $\theta$ be the set of parameters of our model, we will let $\hat{\mu}_{\theta} = \hat{\mu}_{\theta}(x_{n+1}, t_{n+1})$ denote the expectation of $\textbf{X}_n$ predicted by our model and $\overline{\mu}(x_{n+1}, x_0, t_{n+1})$ denote the true expectation of $\textbf{X}_n$ given $\textbf{X}_0 = x_0$ and $\textbf{X}_{n+1} = x_{n+1}$. A brief outline for training this model is as follows:
\vspace{.2cm}
\begin{enumerate}
    \item Apply the forward diffusion process to $x_0$ for a finite number of steps.
    \item For each step, compute $\overline{\mu}(x_{n+1}, x_0, t_{n+1})$ with $\eqref{true_mu_equation}$.
    \item Use the model to predict $\hat{\mu}_{\theta}(x_{n+1}, t_{n+1})$ for each step.
    \item Update $\theta$ to minimize the distance $\left\lVert \overline{\mu} - \hat{\mu}_{\theta} \right\rVert$.
\end{enumerate}
\vspace{.2cm}
With this scheme, the information from $\textbf{X}_0$ gets encoded into the parameter set $\theta$. In other words, our model learns the distribution underlying the original sample.

\subsection{Alternative Methods to Approximate $\overline{\mu}$}
Since our model will know the values of $x_{n+1}$ and $t_{n+1}$, the only unknown quantity in the expression for $\overline{\mu}$ is $x_0$. Thus, instead of predicting $\overline{\mu}$ directly with our model, we may instead opt to predict $x_0$ with our model and then use this prediction to compute an estimate of $\overline{\mu}$. This way, we avoid using the model to predict all the terms which are already known (namely, the terms dependent on $t_{n+1}$ and $x_{n+1}$). 
\\~\\
We also have another method for estimating $\overline{\mu}$.
Using \eqref{eq:Forward_Solution}, we write
    \begin{equation}
        \label{x_0_with_eps}
        x_0 = \frac{x_{n+1} - \beta_{n+1}\eps_0}{\gamma_{n+1}},
    \end{equation}
where $\eps_0$ is sampled from a standard normal distribution.
Substituting this into $\eqref{true_mu_equation}$ and simplifying will yield
\begin{equation}
    \label{mu_with_eps}
    \overline{\mu} = \frac{1}{\gamma(\Delta t_{n+1})}\left(x_{n+1} - \frac{\beta(\Delta t_{n+1})^2}{\beta_{n+1}} \, \eps_0 \right).
\end{equation}
Intuitively, one can think of $\eps_0$ as the "noise" that transforms $x_0$ into $x_{n+1}$. By predicting this noise term $\eps_0$, we can estimate the value of $\overline{\mu}$. Using \eqref{true_mu_equation} and \eqref{mu_with_eps}, we now define two alternative methods for determining $\hat{\mu}_{\theta}$:
\begin{align}
    \label{mu_theta_with_x0}
    \hat{\mu}_{\theta} & = \hat{\mu}_{\theta}(x_{n+1}, t_{n+1}) = \frac{\gamma(\Delta t_{n+1})\beta_n^2}{\beta_{n+1}^2}\cdot x_{n+1} \, + \, \frac{\gamma_n \beta(\Delta t_{n+1})^2}{\beta_{n+1}^2} \cdot (\hat{x_0})_{\theta} \\ \nonumber \\
    \label{mu_theta_with_eps}
    \hat{\mu}_{\theta} & = \hat{\mu}_{\theta}(x_{n+1}, t_{n+1}) = \frac{1}{\gamma(\Delta t_{n+1})}\left(x_{n+1} - \frac{\beta(\Delta t_{n+1})^2}{\beta_{n+1}}\eps_{\theta} \right),
\end{align}
where $(\hat{x_0})_{\theta} = (\hat{x_0})_{\theta}(x_{n+1}, t_{n+1})$ and $\eps_{\theta} = \eps_{\theta}(x_{n+1}, t_{n+1})$ are (respectively) the estimates of $x_0$ and $\eps_0$ produced by our model. To summarize,
we have three methods to estimate $\overline{\mu}$:
\vspace{.2cm}
\begin{enumerate}
    \item Use the model to estimate $\hat{\mu}_{\theta}(x_{n+1}, t_{n+1})$.
    \item Use the model to estimate $(\hat{x_0})_{\theta}$ and compute $\hat{\mu}_{\theta}$ with \eqref{mu_theta_with_x0}.
    \item Use the model to estimate $\eps_{\theta}$ and compute $\hat{\mu}_{\theta}$ with \eqref{mu_theta_with_eps}.
\end{enumerate}
\vspace{.2cm}
Note that methods 2 and 3 can be thought of as equivalent up to a change of variables. However, recall that the unconditional distribution of the $\eps_0$ (i.e., when the initial condition is not known) is standard normal. Thus, the distribution of the $\eps_0$ is potentially simpler than that of the $x_0$ and therefore may be easier to predict. With this idea, we will choose to employ method 3 for estimating $\overline{\mu}$, and the training and generating algorithms discussed in \Cref{sec:train} and \Cref{sec:generate} will use method 3.
\\~\\
With our method for estimating $\overline{\mu}$ determined, we must address how $\theta$ is optimized such that $\hat{\mu}_{\theta}$ approximates $\overline{\mu}$.
\subsection{Learning from Forward Diffusion}
\label{sec:train}
Let $\{x_n\}_{n=1}^{N}$ be a discretized sample path of the forward process starting at $x_0$. Solving for $\eps_0$ in $\eqref{x_0_with_eps}$ gives us
\begin{equation}
    \label{eps_0_exact}
    \eps_0 = \eps_0(x_{n+1}, x_0, t_{n+1}) = \frac{x_{n+1} - \gamma_{n+1} x_0}{\beta_{n+1}}
\end{equation}
for all time steps $n = 0, \ldots, N-1$. In the pursuit of generating new datapoints, we are looking to learn $\eps_0$ without using the initial distribution $\textbf{X}_0$. To achieve this, we will simulate trajectories of the forward process where the initial condition is the sample data. At each step, our model will predict $\eps_{\theta}(x_{n+1}, t_{n+1})$ as an estimate of $\eps_0(x_{n+1}, x_0, t_{n+1})$. Then, the model will be tuned to minimize the distance between $\eps_{\theta}$ and $\eps_0$. In doing so, the model will learn how to predict $\eps_0$ without ever seeing the initial condition. \Cref{training_algo} gives a description of how a diffusion model is trained on a single point $x_0$.
\begin{algorithm}
\caption{Training}\label{training_algo}
\begin{algorithmic}[1]
\STATE \textbf{Parameters:} $x_0 \in \mathbb{R}^d$, $N \in \mathbb{N}$, $0 < t_1 < \ldots < t_N$, $\theta$
\REPEAT
    \STATE Compute $x_1, \ldots, x_N$ using $x_0$ with \eqref{eq:Forward_Solution_Recursive}
    \FOR{$n = 0, \ldots, N-1$}
        \STATE Solve for $\eps_0$ with  \eqref{eps_0_exact}
        \STATE Predict $\eps_{\theta} = \eps_\theta(x_{n+1}, t_{n+1})$ with the model
    \ENDFOR
    \STATE Compute the mean square error (MSE) loss with
    \vspace{-.25cm}
    \begin{equation}
        \label{MSE}
        L[\theta] = \frac{1}{N} \sum_{n=1}^{N} \lvert\lvert \eps_0 - \eps_\theta \rvert\rvert^2
    \end{equation}
    \vspace{-.25cm}
    \STATE Take one gradient descent step on \eqref{MSE}
\UNTIL{converged}
\end{algorithmic}
\end{algorithm}
\subsection{Generating with Reverse Diffusion}
\label{sec:generate}
After training the model on multiple trajectories under the forward process starting at the points in the original sample, we are now able to synthesize arbitrarily many points that "mimic" our original data. \Cref{generating_algo} describes how to generate a new data point. This algorithm simulates the reverse diffusion process -- at each step, $\eps_\theta(x_{n+1}, t_{n+1})$ is computed as an approximation of the true noise $\eps_0(x_{n+1}, x_0, t_{n+1})$. This $\eps_\theta$ is used to estimate $\overline{\mu}$, which then "guides" the reverse trajectories towards an estimate of the initial condition.

\begin{algorithm}
\caption{Generating}\label{generating_algo}
\begin{algorithmic}[1]
    \STATE \textbf{Parameters:} $N \in \mathbb{N}$, $0 < t_1 < \ldots < t_N$, $\theta$
    \STATE Sample $x_N \sim \mathcal{N}(\textbf{0}, \textbf{I})$
    \FOR{$n = N-1, \ldots, 0$}
    \STATE Predict $\eps_\theta(x_{n+1}, t_{n+1})$
    \STATE Compute $\hat{\mu}_\theta = \hat{\mu}_{\theta}(x_{n+1}, t_{n+1})$ with  \eqref{mu_theta_with_eps}
    \STATE Compute $\overline{\sigma}_n$ with  \eqref{std_equation}
    \STATE Sample $z \sim \mathcal{N}(\textbf{0}, \textbf{I})$
    \STATE $x_n = \hat{\mu}_\theta \ + \ \overline{\sigma}_n z $
    \ENDFOR
    \RETURN $x_0$
\end{algorithmic}
\end{algorithm}
\section{Applications for Improving Classifier Performance}
\label{sec:classifier_section}
\begin{figure}[H]
        \centering
        \includegraphics[width=.95\textwidth]{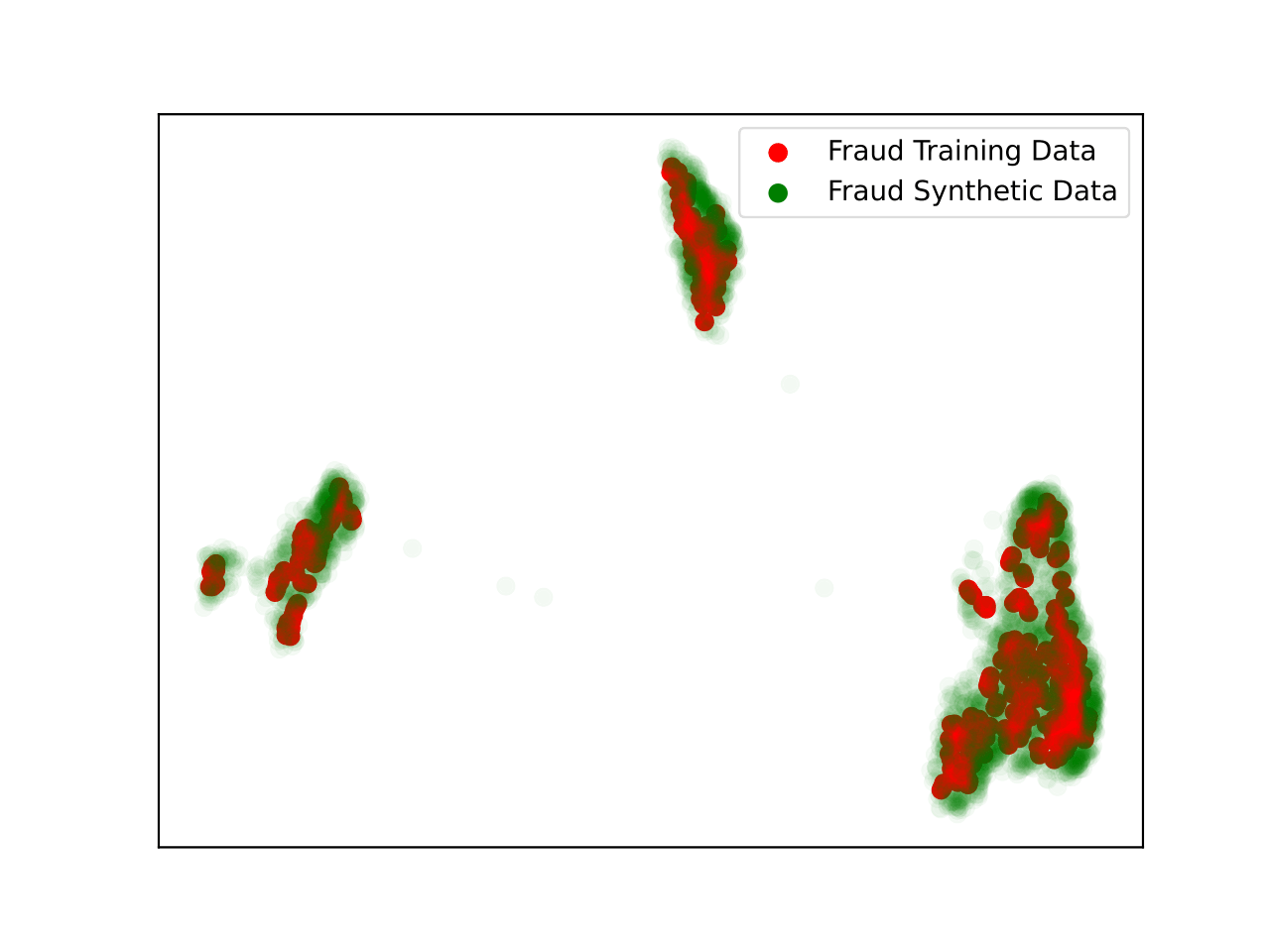}
        \caption{A plot of the training and synthetic fraud data after applying dimensionality reduction with UMAP.}
        \label{fig:UMAP}
\end{figure}

Now, we consider diffusion models as a method to improve classifier performance.
For this, we will use a dataset of credit card transactions \cite{fraud_dataset, fraud_dataset_2} from Kaggle (\href{https://www.kaggle.com/datasets/mlg-ulb/creditcardfraud}{https://www.kaggle.com/datasets/mlg-ulb/creditcardfraud}). This dataset contains information on 284,807 transactions, of which only 492 are fraudulent (this gives a fraud prevalence rate of approximately $1.73 \cdot 10^{-3}$). Because of this, we are interested in generating data that mimics the fraudulent transactions, and then appending this generated data to the training data. In doing so, we may see improvement in a classifier's ability to detect fraud. In particular, we will focus on the precision and recall of a classifier when trained with and without diffusion augmented data, which we define as follows:
\begin{align*}
    \text{Precision} & = \frac{ \text{Number of fraudulent transactions correctly labeled as fraud}}{\text{Total number of transactions labeled as fraud}}, \\~\\
    \text{Recall} & = \frac{ \text{Number of fraudulent transactions correctly labeled as fraud}}{\text{Total number of fraudulent transactions}}. 
\end{align*}
To append synthetic data, we first split the original data into train and test sets. Then, we train a diffusion model on the fraudulent transactions from the training data, generate data with this trained diffusion model, then append this synthetic data to the training data. Because the credit card data has 29 features, we cannot directly visualize the original and synthetic data. Instead, we use dimensionality reduction to obtain a broad idea of the structure of the data. In \Cref{fig:UMAP}, we reduce the dimension of the original and synthetic fraudulent data using Uniform Manifold Approximation and Projection (UMAP). This dimensionality reduction technique leverages theory from algebraic topology to construct a lower-dimensional representation of the data in such a way that preserves topological structure \cite{UMAP}. In this reduced state, the synthetic data mostly agrees with the original fraud data, suggesting that the diffusion model was able to capture the structure underlying the fraudulent transaction data even in 29 dimensions.

\subsection{Classification on the Credit Card Dataset}
\label{sec:classifier_results}
After augmenting the training data with data generated by a diffusion model, we now perform classification. We use the XGBoost \cite{xgboost} and Random Forest \cite{random_forest} classifiers and train them on the unmodified training data and training data augmented with our diffusion models. \Cref{fig:xgb_table} and \Cref{fig:rf_table} show the precision, recall, and $F_1$ score (which is the harmonic mean of precision and recall) of XGBoost and Random Forest applied to the test set after being trained with these two methods. Notably, adding synthetic data improves the recall of both classifiers, but it comes at the tradeoff of precision, meaning both classifiers improve at identifying fraud with the cost of more "false accusations" of fraud.
\\~\\
The GitHub repository containing our implementation of diffusion models, visualization, and classification can be found here: \href{https://github.com/justinle4/Diffusion}{https://github.com/justinle4/Diffusion}. In particular, we utilized the PyTorch \cite{torch} library  in Python to design and train the neural network underlying our diffusion model.

\begin{table}[htb]
    \begin{centering}
    \captionof{table}{XGBoost Results} \label{tab:xgboost} 
    \begin{tabular}{|c|c|c|c|}
    \hline
        Method & Precision & Recall & $F_1$ Score \\ \hline
        No Augmentation & \textbf{0.8901} & 0.8265 & 0.8571 \\ \hline
        Diffusion & 0.8737 & \textbf{0.8469} & \textbf{0.8601} \\ \hline
    \end{tabular}
    \captionsetup{labelformat=empty}
    \caption{Precision, recall, and $F_1$ score for XGBoost on a test set after being trained with and without data augmentation with diffusion models. The maximum value in each column is bolded.}
    \addtocounter{table}{-1}
    \label{fig:xgb_table}
    \end{centering}
\end{table}

\begin{figure}[htb]
    \begin{centering}
    \captionof{table}{Random Forest Results} \label{tab:random_forest} 
    \begin{tabular}{|c|c|c|c|}
    \hline
        Method  & Precision & Recall & $F_1$ Score \\ \hline
        No Augmentation & \textbf{0.9524} & 0.8163 & 0.8791 \\ \hline
        Diffusion & 0.9053 &\textbf{ 0.8776} & \textbf{0.8912} \\ \hline
    \end{tabular}
    \captionsetup{labelformat=empty}
    \caption{Precision, recall, and $F_1$ score for Random Forest on a test set after being trained with and without data augmentation with diffusion models. The maximum value in each column is bolded.}
    \addtocounter{table}{-1}
    \label{fig:rf_table}
    \end{centering}
\end{figure}
\section{Closing Remarks}
In this work, we discussed the mechanics of diffusion models for generating new samples. To do so, we started by defining a forward process as modeled by a linear SDE. From this forward process, a discretized solution to the reverse process could be obtained. The diffusion model is then trained by "witnessing" the forward process (i.e., learning to estimate the noise $\eps_0$), and generates points by inferring the reverse process.
\\~\\
As an application to classification problems in imbalanced datasets, we found that data augmentation with diffusion improved recall on the credit card dataset for two tree-based classifiers while the precision noticeably suffers when augmenting data. That is, models trained on diffusion-augmented data can better detect fraud when it occurs, with the tradeoff of making more "false accusations" of fraud. For classification problems in which failing to detect the minority class is more costly than incorrectly labeling data as belonging to the minority class, data augmentation with diffusion may be advantageous.
\\~\\
For future work on diffusion models, we can consider implementing the Negative Prompting algorithm \cite{negative_prompting}, which currently only sees applications in image generation. The Negative Prompting algorithm allows one to generate data that mimics one class while avoiding another class. On the mathematical front, connections between the reverse SDE in \cite{thiery} and partial differential equations can be investigated utilizing a result known as Itô’s formula (discussed in \cite{evans}). Such an approach would

\appendix

\section{Solving the Ornstein-Uhlenbeck Equation}
\label{sec:forward_solution}
Solving \eqref{eq:Ornstein-Uhlenbeck} and SDEs more broadly requires the development of a new type of integration called  "Itô integration". This is what allows us to make sense of integrating against $\mathrm{d}\textbf{B}_t$, which, unlike the traditional notions of integration, integrates with respect to a stochastic process. A rigorous development of the Itô integral and the necessary properties of the integral to obtain  \eqref{eq:Ornstein-Uhlenbeck_Solution_Unsimplified} can be found in Evans' text on SDE \cite{evans}. The key lemma from this text is that integrating against $\mathrm{d}\textbf{B}_t$ does not yield a number, but rather a random variable.
\begin{lemma}
    \label{Evans_Lemma}
    Let $0 \leq a < b$ and $g \in \mathbb{L}^2(a, b)$. Then, $\int_{a}^b g(s) \, \mathrm{d}\mathbf{B}_s$ is normally distributed with
    \begin{align*}
        \mathbb{E}\left[\int_{a}^{b} g(s) \, \mathrm{d}\mathbf{B}_s \right] = 0
    \end{align*}
    and
    \begin{align*}
        \mathbb{E}\left[\left(\int_{a}^{b} g(s) \, \mathrm{d}\mathbf{B}_s \right)^2\right] = \int_{a}^{b} g^2(s) \, ds.
    \end{align*}
\end{lemma}
The proof of this lemma can also be found in \cite{evans} and is one of the key properties of the Itô integral. With this lemma, \eqref{eq:Ornstein-Uhlenbeck_Solution_Unsimplified} can easily be obtained.
\begin{proof} \textbf{(\Cref{Prop_OUE_solution})}
    Our first step towards solving \eqref{eq:Ornstein-Uhlenbeck} is to make use of the method of integrating factors. Set $\textbf{Z}_t = e^t\textbf{X}_t$ and observe that
    \begin{align*}
        \mathrm{d}\textbf{Z}_t & = e^t\textbf{X}_t \, \mathrm{d}t + e^t\mathrm{d}\textbf{X}_t = e^t\textbf{X}_t \, \mathrm{d}t + e^t(-\textbf{X}_t \, \mathrm{d}t + \sqrt{2}\mathrm{d}\textbf{B}_t) \\
        & = \sqrt{2}e^t\mathrm{d}\textbf{B}_t.
    \end{align*}
    Integrating both sides over $[0, t]$ then yields
    \begin{align*}
         \textbf{Z}_t = \textbf{Z}_0 + \sqrt{2}\int_0^t e^s \, \mathrm{d}\textbf{B}_s
        \implies \textbf{X}_t = e^{-t}\textbf{X}_0 + \sqrt{2}\int_0^t e^{-(t-s)} \, \mathrm{d}\textbf{B}_s,
    \end{align*}
    establishing \eqref{eq:Ornstein-Uhlenbeck_Solution_Integral}.
    From here, we use \Cref{Evans_Lemma} to deduce
    \begin{align*}
        \mathbb{E}\left[\sqrt{2}\int_0^t e^{-(t-s)} \, \mathrm{d}\textbf{B}_s \right] = (\sqrt{2}e^{-t}) \mathbb{E}\left[\int_0^t e^s \, \mathrm{d}\textbf{B}_s \right] = 0
    \end{align*}
    and
    \begin{align*}
    \mathbb{E}\left[\left(\sqrt{2}\int_0^t e^{-(t-s)} \, \mathrm{d}\textbf{B}_s\right)^2 \right] & = (2e^{-2t}) \mathbb{E}\left[\left(\int_0^t e^s \, \mathrm{d}\textbf{B}_s\right)^2 \right] = (2e^{-2t}) \int_0^{t} e^{2s} \, ds = e^{-2(t-s)}\Bigg\rvert_{s=0}^{t} \\
        & = 1 - e^{-2t}.
    \end{align*}
    Now, we conclude
    \begin{align*}
        \text{Var}\left[\sqrt{2}e^{-t}\int_0^t e^s \, \mathrm{d}\textbf{B}_s \right] & = \mathbb{E}\left[\left(\sqrt{2}e^{-t}\int_0^t e^s \, \mathrm{d}\textbf{B}_s\right)^2 \right] - \mathbb{E}\left[\sqrt{2}e^{-t}\int_0^t e^s \, \mathrm{d}\textbf{B}_s \right]^2 \\
        & = 1 - e^{-2t}.
    \end{align*}
   In other words,
   \begin{align*}
       \sqrt{2}e^{-t} \int_0^t e^s \, \mathrm{d}\textbf{B}_s \sim \mathcal{N}(\textbf{0}, 1-e^{-2t}),   
    \end{align*}
    so we can rewrite $\textbf{X}_t$ as follows:
    \begin{align*}
        \textbf{X}_t = e^{-t}\textbf{X}_0 + \sqrt{1 - e^{-2t}} \, \textbf{Z}
    \end{align*}
    with $\textbf{Z} \sim \mathcal{N}(0, \textbf{I})$. This is \eqref{eq:Ornstein-Uhlenbeck_Solution_Unsimplified}.
\end{proof}
To obtain \eqref{eq:Forward_Solution_Recursive}, we simply repeat the steps above, but instead integrating over $[t_n, t_{n+1}]$ rather than $[0, t]$.

\section{Time Discretization}
\label{sec:time_discretization}
Here, we describe an algorithm to choose the values of $\Delta t_n$. Importantly, we do not explicitly set the values of $\Delta t_n$; rather, we control the values of $\beta_n$, where the $\beta_n$ are as defined in \eqref{eq:beta_def}. This is because $\beta_n^2$ is the variance of $\mathbf{X}_n$ at each time step, and having control over the variability of $\mathbf{X}_n$ is more immediately interpretable than the size of a time step (that is, we do not know a priori what makes a "good" choice for a time step size). These $\beta_n$ are chosen to be equally spaced between the user-defined values $\beta_1$ and $\beta_N$, and $\Delta t_n$ is then obtained using $\beta_n$. With this scheme, $\Delta t_n$ increases with respect to $n$, meaning more time steps will be found closer to 0 (i.e., when the "drift" portion of forward diffusion is most pronounced). Hence, a model trained with such a time discretization scheme will more thoroughly learn the drift, which is crucial when generating points with the reverse process. In contrast, diffusion models trained using uniform time steps tend to produce synthetic data that is less faithful to the original sample since more weight is placed on learning the forward process for large $t$. This is the stage of forward diffusion when the data has already become noisy and remains noisy, so learning these latter parts of the forward process are not as important for inferring the reverse process.
\begin{algorithm}
\caption{Time Discretization}\label{discretization_algo}
\begin{algorithmic}[1]
    \STATE \textbf{Parameters:} $N \in \mathbb{N}$, $0 < \beta_1 < \beta_N < 1$.
    \FOR{$n = 1, \ldots, N$}
        \STATE $\beta_n = \beta_1 + \cfrac{\beta_N - \beta_1}{N-1} \cdot (n-1)$
        \STATE $\Delta t_n = -\cfrac{1}{2}\ln(1 - \beta_n)$
        \STATE $t_n = \displaystyle {\sum\limits_{k=1}^{n}} \, \Delta t_k$
    \ENDFOR
    \RETURN $(t_1, \ldots, t_N), (\Delta t_1, \ldots, \Delta t_N)$
\end{algorithmic}
\end{algorithm}

\FloatBarrier

\section{Deriving the Discretized Reverse Diffusion}
\label{sec:reverse_proof}
We present a proof to \Cref{prop:reverse_prop}, which crucially leverages our discrete time steps by using Bayes' theorem.
\begin{proof}\textbf{(\Cref{prop:reverse_prop})}
    First, Bayes' theorem gives us
    \begin{equation*}
    \begin{aligned}[t]
        \rho(x_n \, \vert \, x_{n+1}, x_0) = \frac{\rho(x_{n+1} \, \vert \, x_n, x_0) \rho(x_n \, \vert \, x_0)}{\rho(x_{n+1} \, \vert \, x_0)}.
    \end{aligned}
    \end{equation*}
    Note that by \eqref{eq:Forward_Solution_Recursive}, the $\textbf{X}_n$ all have the Markov property (in other words, the position of a trajectory at one time step depends only on its position at the previous time step). Hence, $\rho(x_{n+1} \, \vert \, x_n, x_0) = \rho(x_{n+1} \, \vert \, x_n)$. Moreover, $x_{n+1}$ and $x_0$ are fixed, so the joint density $\rho(x_{n+1} \, \vert \, x_0)$ is a constant. So, we have $\rho(x_{n+1} \, \vert \, x_n) \sim \mathcal{N}(\gamma(\Delta t_{n+1})x_n, \beta(\Delta t_{n+1})^2)$ by \Cref{cor_recursive} and $\rho(x_{n} \, \vert \, x_0) \sim \mathcal{N}(\gamma_n x_0, \beta_n^2)$ by \Cref{Prop_OUE_solution}. Since we already know that $\rho(x_n \, \vert \, x_{n+1}, x_0)$ is a density, we only need to consider the kernel of the right hand side, meaning all constants can be ignored. In particular, we will take all terms that are not dependent on $x_n$ and collect them into constants that we will denote as $C_1$ and $C_2$:
    \scriptsize
    \begin{align*}
        \displaystyle{}
        \rho(x_n \, \vert \, x_{n+1}, x_0) & = \frac{\rho(x_{n+1} \, \vert \, x_n) \rho(x_n \, \vert \, x_0)}{\rho(x_{n+1} \, \vert \, x_0)} \\
        & \propto \rho(x_{n+1} \, \vert \, x_n) \rho(x_n \, \vert \, x_0) \\
        & \propto \exp\left(-\frac{1}{2\beta(\Delta t_{n+1})^2}\cdot [x_{n+1} - \gamma(\Delta t_{n+1}) x_n]^2\right) \cdot \exp\left(-\frac{1}{2\beta_n^2}\cdot [x_{n} - \gamma_n x_0]^2\right) \\
        & = \exp\left(-\frac{x_{n+1}^2 - 2\gamma(\Delta t_{n+1})x_{n+1}x_n + \gamma(\Delta t_{n+1})^2 x_n^2}{2\beta(\Delta t_{n+1})^2}\right) \cdot \exp\left(-\frac{x_{n}^2 - 2\gamma_n x_nx_0 + \gamma_n^2x_0^2}{2\beta_n^2}\right) \\
        & = \exp\left(-\frac{\beta_n^2(x_{n+1}^2 - 2\gamma(\Delta t_{n+1})x_{n+1}x_n + \gamma(\Delta t_{n+1})^2x_n^2) + \beta(\Delta t_{n+1})^2(x_{n}^2 - 2\gamma_n x_nx_0 + \gamma_n^2x_0^2)}{2\beta(\Delta t_{n+1})^2\beta_n^2} \right) \\
        & = \exp\left(-\frac{[\beta_n^2\gamma(\Delta t_{n+1})^2 + \beta(\Delta t_{n+1})^2]x_n^2 - 2[\gamma(\Delta t_{n+1})\beta_n^2 x_{n+1} + \gamma_n \beta(\Delta t_{n+1})^2x_0]x_n + C_1}{2\beta(\Delta t_{n+1})^2\beta_n^2} \right).
    \end{align*}
    \normalsize
    Notice that
    \begin{align*}
        \beta_n^2\gamma(\Delta t_{n+1})^2 + \beta(\Delta t_{n+1})^2 & = (1-e^{-2t_n})e^{-2\Delta t_n} + (1 - e^{-2\Delta t_n}) \\
        & = e^{-2\Delta t_n} - e^{-2t_{n+1}} + (1 - e^{-2\Delta t_n}) \\
        & = 1 - e^{-2t_{n+1}} \\
        & = \beta_{n+1}^2.
    \end{align*}
    Using this fact, we find
    \small
    \begin{align*}
        \rho(x_n \, \vert \, x_{n+1}, x_0) & \propto \exp\left(-\frac{\beta_{n+1}^2x_n^2 - 2[\gamma(\Delta t_{n+1})\beta_n^2 x_{n+1} + \gamma_n \beta(\Delta t_{n+1})^2x_0]x_n}{2\beta(\Delta t_{n+1})^2\beta_n^2} \right) \cdot \exp\left(\frac{C_1}{2\beta(\Delta t_{n+1})^2\beta_n^2}\right) \\
        & \propto \exp\left(-\frac{\beta_{n+1}^2x_n^2 - 2[\gamma(\Delta t_{n+1})\beta_n^2 x_{n+1} + \gamma_n \beta(\Delta t_{n+1})^2x_0]x_n}{2\beta(\Delta t_{n+1})^2\beta_n^2} \right) \\
        & \propto \exp\left(-\frac{x_n^2 - 2 \cdot \frac{\gamma(\Delta t_{n+1})\beta_n^2 x_{n+1}}{\beta_{n+1}^2}x_n - 2 \cdot \frac{\gamma_n \beta(\Delta t_{n+1})^2x_0}{\beta_{n+1}^2}x_n}{2 \cdot \frac{\beta(\Delta t_{n+1})^2\beta_n^2}{\beta_{n+1}^2}} \right) \\
        & = \exp\left(-\frac{\left(x_n - \frac{\gamma(\Delta t_{n+1})\beta_n^2 x_{n+1}}{\beta_{n+1}^2} - \frac{\gamma_n \beta(\Delta t_{n+1})^2x_0}{\beta_{n+1}^2}\right)^2 + C_2}{2 \cdot \frac{\beta(\Delta t_{n+1})^2\beta_n^2}{\beta_{n+1}^2}} \right) \\
        & \propto \exp\left(-\frac{\left(x_n - \left(\frac{\gamma(\Delta t_{n+1})\beta_n^2}{\beta_{n+1}^2} \cdot x_{n+1} + \frac{\gamma_n \beta(\Delta t_{n+1})^2}{\beta_{n+1}^2} \cdot x_0\right)\right)^2}{2 \cdot \frac{\beta(\Delta t_{n+1})^2\beta_n^2}{\beta_{n+1}^2}} \right).
    \end{align*}
    \normalsize
    We recognize this as the kernel of a $\mathcal{N}(\overline{\mu}, \overline{\sigma}^2_n)$ distribution.
\end{proof}

\section*{Acknowledgments}
This article began as my honors thesis for my Bachelor's degree at Arizona State University's School of Mathematical and Statistical Sciences. I would like to thank my advisors, Dr.\ Sebastien Motsch and Dr.\ Johannes Brust, for their mentorship, expertise, and inspiration. Thank you as well to the School of Mathematical and Statistical Sciences and Barrett, the Honors College at Arizona State University for giving me the support to succeed as an undergrad. Finally, thank you to my high school calculus teacher Mr.\ Tutt, who ignited my passion for mathematics and started me on this incredible journey.

\bibliography{references}

\end{document}